\documentclass[10pt, a4paper]{article}

\usepackage{lrec-coling2024} 

\usepackage{adjustbox}      
\usepackage{graphicx}
\usepackage{makecell}
\usepackage{lineno}
\usepackage{booktabs}
\usepackage{amsfonts}
\usepackage{amssymb}
\usepackage{bbm}
\usepackage{amsmath}
\usepackage{color}
\usepackage{CJKutf8}
\newcommand{\nop}[1]{}

\usepackage{marvosym}

\title{Negation Triplet Extraction with Syntactic Dependency\\ and Semantic Consistency}

\name{\begin{tabular}{c}
Yuchen Shi$^{1\dag}$, Deqing Yang$^{1\ddag\textrm{\Letter}}$, 
Jingping Liu$^{2\S}$, Yanghua Xiao$^{1\ddag}$\\ Zongyu Wang$^{3\P}$, Huimin Xu$^{3\P}$
\end{tabular}}
\address{
$^1$School of Data Science, Fudan University, Shanghai Key Laboratory of Data Science, Shanghai, China\\
$^2$East China University of Science and Technology, Shanghai, China\\
$^3$Meituan, China\\
$^\dag$\texttt{ycshi21@m.fudan.edu.cn}, $^
\ddag$\texttt{\{shawyh,yangdeqing\}@fudan.edu.cn}\\
$^\S$\texttt{jingpingliu@ecust.edu.cn}, 
$^\P$\texttt{\{zongyu.wang, huimin.xu\}@meituan.com}\\
}

\abstract{
Previous works of negation understanding mainly focus on negation cue detection and scope resolution, without identifying negation subject which is also significant to the downstream tasks. In this paper, we propose a new negation triplet extraction (NTE) task which aims to extract negation subject along with negation cue and scope. To achieve NTE, we devise a novel \textbf{S}yntax\&\textbf{S}emantic-\textbf{E}nhanced \textbf{N}egation \textbf{E}xtraction model, namely \textbf{SSENE}, which is built based on a generative pretrained language model (PLM) {of Encoder-Decoder architecture} with a multi-task learning framework. Specifically, the given sentence's syntactic dependency tree is incorporated into the PLM's encoder to discover the correlations between the negation subject, cue and scope. Moreover, the semantic consistency between the sentence and the extracted triplet is ensured by an auxiliary task learning. Furthermore, we have constructed a high-quality Chinese dataset \textbf{NegComment} based on the users' reviews from the real-world platform of Meituan, upon which our evaluations show that SSENE achieves the best NTE performance compared to the baselines. Our ablation and case studies also demonstrate that incorporating the syntactic information helps the PLM's recognize the distant dependency between the subject and cue, and the auxiliary task learning is helpful to extract the negation triplets with more semantic consistency. 
We further demonstrate that SSENE is also competitive on the traditional CDSR task. 
 \\ \newline  }

\begin{document}

\maketitleabstract

\begin{CJK}{UTF8}{gbsn}

\section{Introduction}
Negation is a common linguistic phenomenon in natural language. Negation understanding has become an important task in natural language understanding, which is crucial to many downstream tasks such as sentiment analysis, question answering, Web search and natural language inference \cite{Kassner2020}. 
For example, Table \ref{tb:example} displays some users' review sentences from the famous Chinese service platform \emph{Meituan}\footnote{\url{https://www.meituan.com/}.}, each of which includes the negation expressions. For Example 1 in the table, if the user's negation expression (soundproof is not good) is correctly recognized by the system, the hotel would be filtered out from Meituan's recommendation list for a user who very cares about the soundproof effect. 

\begin{table*}[t]
\centering
\scriptsize{
\begin{tabular}{p{0.05\textwidth}|p{.4\textwidth}|p{.4\textwidth}}
    \toprule
   ID & Sentence example & Negation triplet \\
    \hline
    1& \#隔音不太好\#门外声音很清楚。\#Soundproof is not good\#Sounds outside the door are very clear. & <隔音Soundproof, 不not, 好good> \\ 
    \hline
    2& 四楼，胡安的画作，小鸟女人星星，这个是符号化的画作，不太好懂。 On the fourth floor, Juan's painting, Little Bird Woman Star, this is a symbolic painting, and not easy to understand. & <小鸟女人星星Little Bird Woman Star, 不not, 好懂easy to understand> \\ 
    \hline
    3& 酒店的早餐不好吃，食材也不新鲜，没有多少可选择，完全不符合星级酒店的标准。 The breakfast at the hotel is not tasty, the ingredients are not fresh, not much to choose from, and it completely does not meet the standards of a star-rated hotel. & <早餐breakfast, 不not, 好吃tasty>, <食材ingredients, 不not, 新鲜fresh>, <早餐breakfast, 没有not, 可选择much to choose from>, <早餐breakfast, 不not, 符合星级酒店的标准meet the standards of a star-rated hotel> \\
    \bottomrule
\end{tabular}
}
\caption{Some toy examples of negation triplet extraction from Meituan's user reviews. The original Chinese sentences and spans are translated into English ones.}\label{tb:example}
\end{table*}

Up to now, some efforts \cite{truong2022improving,peng2018negbio,Focus1,Focus2,Focus3} have been made for negation understanding, especially for cue detection and scope resolution (CDSR) \cite{truong2022improving}. The former is detecting the phrase triggering the negation, and the latter is determining the affected span that is negated. Inspired by the power of pretrained language models (PLMs) on many natural language processing (NLP) tasks, PLMs were also employed to achieve CDSR \cite{lockard2020web}. 
Although these models have demonstrated effectiveness on CDSR, they neglect extracting the negation subjects which are critical to many downstream applications. Recall Example 1 in Table \ref{tb:example}, the negation cue is ``not'' and the scope is ``good'', while the negation subject is ``soundproof''. If this subject is not identified, we cannot align the attribute of the hotel that is negated by the user, which will degrade the precise hotel recommendation. 

To address this critical problem, in this paper we focus on a new negation extraction task which is formalized as extracting the negation triplet(s) of form \emph{<subject, cue, scope>} from a given sentence. Such triplets sufficiently describe the attributes of the entity mentioned in the sentence that are negated, as shown in the toy examples in Table \ref{tb:example}. 
Compared with previous CDSR task, the negation triplets in our proposed extraction task have the following particularities which pose some challenges for the solutions based on PLMs: (1) A negation subject may appear at the position distant to the cue and scope in the sentence, introducing long-distance dependency issue. This requires the model to understand the overall syntactic structure of the sentence for accurate triplet extraction. For example, in Example 2 of Table \ref{tb:example}, ``Little Bird Woman Star'' and ``Juan's painting'' are appositive phrases, and both distant to ``not easy to understand''. (2) In a sentence of real-world user review, there may exist multiple diverse entities and negation relations. For example, in Example 3 of Table \ref{tb:example}, there are several entities that may be the negation subjects, along with multiple cues and scopes. They might be mistakenly paired by the model, resulting in the semantics inconsistent with the sentence's semantics (refer to the case study in our experiments).

\begin{figure}[t]
    \centering
    \includegraphics [width=1.1\columnwidth]{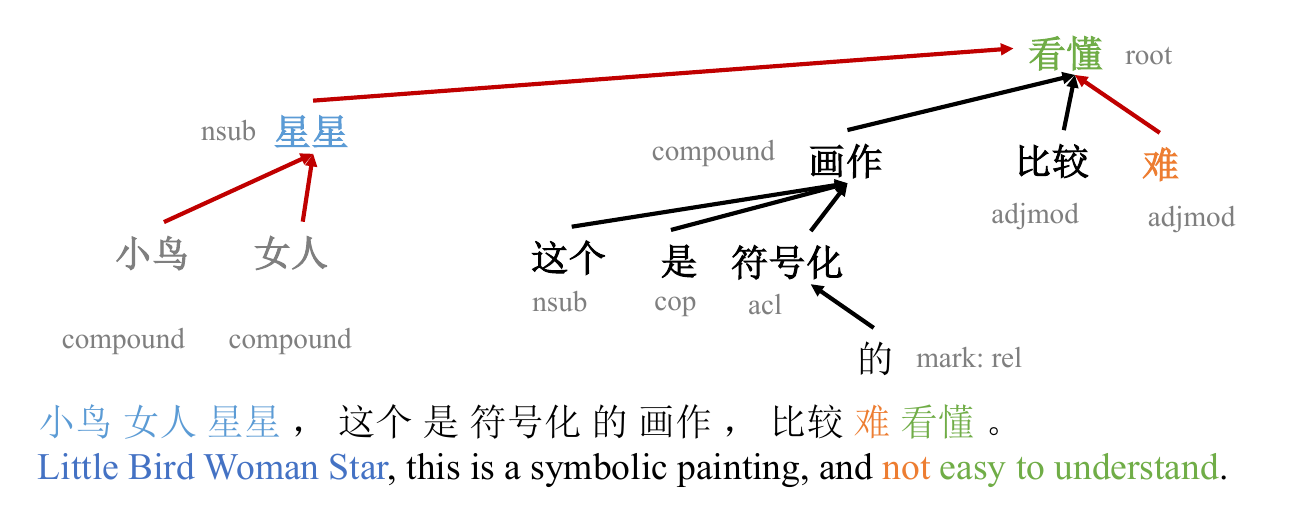} %
    \vspace{-.8cm}
    \caption{An examples of negation triplet extraction with syntactic dependency tree (better viewed in color). The negation subject, cue and scope are marked with blue, orange and green, respectively.}
    \label{fig:data}
\end{figure}

Facing these challenges, in this paper we devise a novel \textbf{S}yntax \& \textbf{S}emantic-\textbf{E}nhanced \textbf{N}egation \textbf{E}xtraction model named as \textbf{SSENE}, to achieve effective negation triplet extraction (NTE). 
SSENE enhances the Encoder-Decoder architecture from both syntax and semantic perspectives, incorporating the following two main innovations: 
(1) Within SSENE, the \emph{syntax-aware encoder} integrates the sentence's syntactic information alongside its semantic information, thereby crafting a better representation of the sentence. 
As shown in Fig. \ref{fig:data}, the sentence's syntactic dependency tree well indicates the correlations between the negation subject, cue and scope, which is helpful to achieve accurate NTE. (2) SSENE adopts a multi-task learning framework, where an auxiliary task is introduced to ensure the semantic consistency between the sentence and the extracted negation triplet(s). 

Moreover, to fine-tune and evaluate the models' performance on NTE, we have manually constructed a new Chinese dataset \textbf{NegComment} which was collected from the massive user reviews in Meituan's platform of daily life. 
To ensure NegComment's high quality, we also designed a well-tailored annotation process that will be introduced in the experiment section.
In summary, our contributions include:
    
     1. We propose a new extraction task NTE, which is an important task of negation understanding and valuable to downstream applications.
    
     2. For NTE, we propose a novel model SSENE based on the Encoder-Decoder architecture with a multi-task learning framework, which leverages the sentence's syntactic dependency information and the semantic consistency between the extracted triplets and the sentence, to obtain enhanced performance.
    
     3. We have constructed a high-quality dataset NegComment from the real-world scenarios, which can be used to fine-tune and evaluate the models of negation understanding.
    
     4. We have conducted extensive experiments to demonstrate that {our proposed SSENE outperforms the state-of-the-art (SOTA) baselines including ChatGPT}, and justify the impacts of incorporating the syntactic dependency and semantic consistency. {Our further studies verify that SSENE also performs well on traditional CDSR task. }

\section{Methodology}
\subsection{Task Formalization} \label{sec:task-defination}
The NTE task we proposed in this paper can be formalized as follows. 
Given an input sentence $x=\left[w_1, \ldots, w_n\right]$, where $w_i(1\leq i\leq n)$ is the $i$-th token, the objective of NTE is to extract a series of negation triplets $T=\{t_1, t_2, ..., t_{|T|}\}=\{<s_i, p_i, o_i>\}_{i=1}^{|T|}$ that express negation semantics in $x$. 
Here, $s_i, p_i$ and $o_i$ correspond to the negation subject, cue and scope in negation triplet $t_i$, respectively. In general, $s_i$ and $o_i$ could be either a term or a phrase, or even a clause in $x$, which is the major particularity different to the generic triplet of <head\_entity, relation, tail\_entity>. 
Similar to the constraints in \cite{sfu_lr}, we require that each element of the negation triplet is a continuous span in the sentence, and there is no overlap between different elements.

\begin{figure*}[t]
    \centering
    \includegraphics[width=\textwidth]{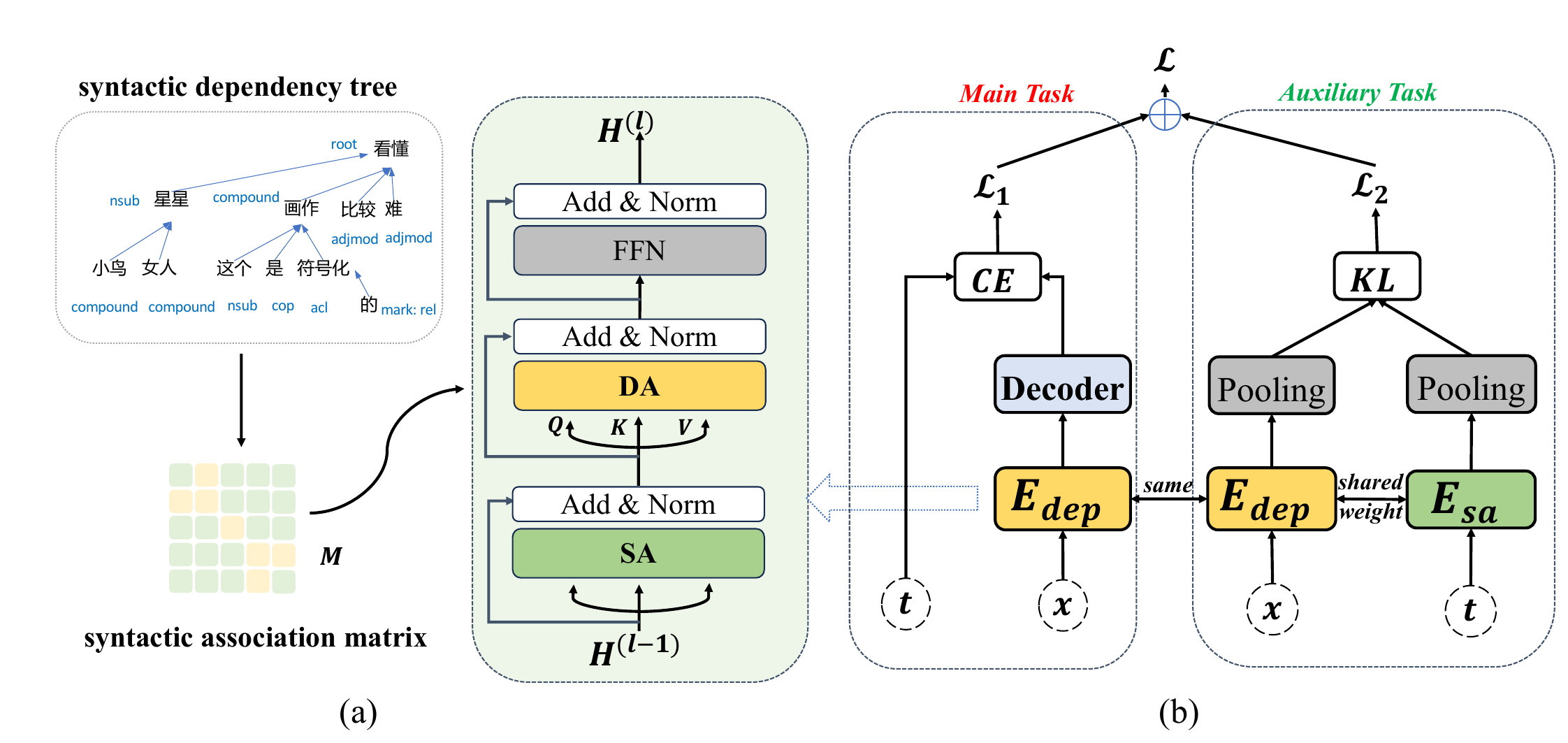} 
    \caption{Subfigure (b) displays the overall framework of SSENE, in which the encoder $\mathbf{E}_{dep}$ is depicted in Subfigure (a). 
    }
    \label{fig:framework}
\end{figure*}

\subsection{Model Overview} \label{sec:model}
The overall architecture of our proposed SSENE is shown in Fig. \ref{fig:framework} (b), which adopts a multi-task learning framework detailed in Section \ref{sec:arch}. The primary task of SSENE is just NTE, where we use the Encoder-Decoder PLMs to generate the extraction results, and the auxiliary task is proposed to enhance the semantic consistency of the extracted triples with the original sentence. 
Moreover, the sentence' syntactic information has been proven significant in negative extraction tasks~\cite{Zou2015,mckenna2019learning,mahany2022negation}, so we explicitly introduce the syntactic information into our model, which is implemented in the syntax-aware encoder $\mathbf{E}_{dep}$ as shown in \ref{fig:framework} (a). We will introduce the details of $\mathbf{E}_{dep}$ in Section \ref{sec:layer-dep}.

\subsection{Multi-task Learning Framework}\label{sec:arch}
We first introduce the multi-task learning framework in SSENE. 
For the main task in SSENE learning, we adopt the Encoder-Decoder architecture to generate the negation triplets given the input sentence $x$, where the targeted triplets $T=\{<s_i, p_i, o_i>\}_{i=1}^{|T|}$ are serialized in the form $y = s_1[S]p_1[S]o_1[SEQ]...[SEQ]s_T[S]p_T[S]o_T$ (the special token $[S]$ and $[SEQ]$ are used to separate the elements within a triplet and across different triplets, respectively), and served as the target output of the generative model. 
In the encoder, as we mentioned, $x$'s syntactic information is useful for extracting the negation triplets, so we specifically design a syntax-aware encoder $\mathbf{E}_{dep}$ to incorporate syntactic information into $x$'s representation. In contrast to a normal self-attention encoder (denoted as $\mathbf{E}_{sa}$),  $\mathbf{E}_{dep}$ contains both a self-attention layer (SA) and a dependency-attention layer (DA), which feeds the sentence's dependency structure into the encoder by transforming it into a \emph{syntactic association matrix}. The detailed operations in $\mathbf{E}_{dep}$ will be described in Section~\ref{sec:layer-dep}. 
Then, the decoder is used to generate the extracted negation triplet based on $x$'s representation. 
As other generative PLMs \cite{vaswani2017attention,dong2019unified,zhou2020pre}, given a real triplet $t$, we adopt the following cross-entropy (CE) as main task's loss,
\begin{equation}
\mathcal{L}_1 = -\frac{1}{N}\sum_{i=1}^{N}\sum_{j=1}^{M}y_i^j\log(\hat{y}_i^j),
\end{equation}
where $N$ is the length of $t$'s token sequence, and $M$ is the vocabulary size. $y_i^j=1$ if the $i$-th token in $t$ is just the $j$-th token in the vocabulary, otherwise $y_i^j=0$. And $\hat{y}_i^j$ is the probability that the $j$-th token in the vocabulary is predicted by the model as the $i$-th token in $t$.

For a correct negation triplet extracted from a given sentence, it should not express the inconsistent semantics w.r.t. the sentence. Inspired by this intuition, we propose an auxiliary task learning in our SSENE, which encourages the extracted negation triplet's semantics to be consistent with the sentence's semantics. 
Specifically, in the auxiliary task, besides using $\mathbf{E}_{dep}$ to obtain the input $x$'s representation, we further use a basic self-attention encoder $\mathbf{E}_{ sa}$ to obtain the representation of the targeted negation triplet $t$. 
In order to conveniently compute the semantic distance between the representations of $x$ and $t$, we further adopt a pooling layer to transfer their representations into two embeddings of $d$ dimensions, which are denoted as $\mathbf{x}$ and $\mathbf{t}$, respectively. Then, the KL divergence is used as the auxiliary task's loss,
\begin{equation}
\mathcal{L}_2 = \mathbb{D}_{KL}(\mathbf{t} \| \mathbf{x}) 
= \sum_{i=1}^d {\mathbf{t}^i} \ln \left(\frac{\mathbf{t}^i}{\mathbf{x}^i}\right),
\end{equation}
where $\mathbf{t}^i$ ($\mathbf{x}^i$) is the $i$-th element of embedding $\mathbf{t}$ ($\mathbf{x}$). Notice that $\mathbf{E}_{dep}$ and $\mathbf{E}_{sa}$ share the same weight parameters, except that $\mathbf{E}_{dep}$ has a DA layer to additionally leverage the syntactic information (Section \ref{sec:layer-dep}). The triplet $t$ is only used as the input of the auxiliary task during training, whose goal is to optimize the shared weights of the two encoders. In validation phase, we only use $\mathbf{E}_{dep}$ to encode the input sentence $x$ and obtain the predicted triplets. 

Thus, the overall loss of SSENE is
\begin{equation}\label{eq:L}
\mathcal{L} = \mathcal{L}_1 + \alpha \mathcal{L}_2,
\end{equation}
where $0<\alpha<1$ is the controlling parameter. We use joint training strategy for SSENE's multi-task learning.

\subsection{Syntax-aware Encoder} \label{sec:layer-dep}
In this subsection, we detail the syntax-aware encoder in SSENE, denoted as $\mathbf{E}_{dep}$, which we specially designed for the goal of incorporating the sentence's syntactic information.

$\mathbf{E}_{dep}$ is built based on the vanilla Transformer encoder \cite{vaswani2017attention}. 
Given that syntactic structure aids in negation understanding\cite{mckenna2019learning}, we introduce a Dependency-Attention layer(DA) in each $\mathbf{E}_{dep}$, in addition to Self-Attention layer(SA) and a feedforward network (FFN), as depicted in Figure \ref{fig:framework} (a).
Overall, to incorporate $x$'s syntactic information, we first get the syntactic dependency tree of $x$ through a syntax parser, then use the distance information on the tree to measure the relationships between the tokens in $x$, which is finally transformed and introduced into the DA layer in a manner similar to a mask matrix \cite{vaswani2017attention} as follows.

Specifically, we compute a \emph{syntactic association matrix} $\mathbf{M}\in\mathbb{R}^{n\times n}$ for the input sentence $x$, where $n$ is the token number of $x$. Each element $m_{ij} (1\leq i,j\leq n)$ of $\mathbf{M}$ is a syntactic association score quantifying the syntactic dependency importance between token $w_i$ and $w_j$ in $x$. We compute $m_{ij}$ based on the distance between $w_i$ and $w_j$ on $x$'s syntactic dependency tree as below,
\begin{equation}\label{eq:M}
m_{ij} = \operatorname{Softmax}\big(f(d_{ij})\big),
\end{equation}
where $d_{ij}$ is the distance between $w_i$ and $w_j$ computed as \cite{ahmad2021gate}, {Softmax is calculated by rows}. The convert function $f()$ is a monotonic decreasing functions of $d_{ij}$, here we choose $f(x) = \frac{\gamma_1}{\gamma_2 + x}$.

To incorporate the information of the syntactic association matrix $\mathbf{M}$ into the self-attention mechanaism, we refine the self-attention's computation as\footnote{Note that since $\mathbf{M}$ has been normalized by the Softmax operation in Eq. \ref{eq:M}, we remove the scaling factor $\sqrt{d}$ of standard self-attention.},
\begin{equation}\label{eq:A}
\mathbf{A}_{dep} = \operatorname{Softmax}\left(\mathbf{Q}\mathbf{K}^T \odot \mathbf{M}\right)\in\mathbb{R}^{n\times n},
\end{equation}
where $\mathbf{Q},\mathbf{K}\in\mathbb{R}^{n\times d}$ are the query and key matrix, respectively. And $\odot$ denotes element-wise product. 
Finally, the output of a DA sublayer is
\begin{equation}
\Tilde{\mathbf{H}} = \mathbf{A}_{dep} \mathbf{V}= \operatorname{Softmax}\left(\mathbf{Q}\mathbf{K}^T \odot \mathbf{M}\right)\mathbf{V}\in\mathbb{R}^{n\times d},
\end{equation}
where $\mathbf{V}\in\mathbb{R}^{n\times d}$ is the value matrix.

\section{Experiments}
In this section, we display the results of extensive experiments to answer the following research questions\footnote{Our dataset, source code of SSENE, and the designed prompts for ChatGPT are available on \url{https://github.com/Easonsi/SSENE}.}: 

RQ1: Can our proposed SSENE achieve superior performance over the other models on NTE?

RQ2: Are the two major designs in SSENE useful to performance gains, i.e., incorporating the sentence's syntactic dependency information and the auxiliary task of semantic consistency?

RQ3: Is the way of incorporating the syntactic dependency information into SSENE's syntax-aware encoder more effective than other ways?

{RQ4: Can the two major designs in SSENE be generalised to other negation understanding tasks such as CDSR?}

\subsection{Dataset Construction} \label{sec:data}
As we mentioned before, the data of NegComment used to evaluate the models' performance on NTE was collected from massive user reviews on the platform of Meituan, which is one of Chinese largest online service platforms for daily life. 
From the candidate user reviews, we selected the sentences containing negation expressions related to seven domains: hotel, tourism, restaurant, leisure and entertainment, sports and fitness, medical care, and beauty.

To obtain more reliable annotated data, we formulated the annotation guidelines in advance, according to the relevant rules from CDSR \cite{sfu_lr,bioscope_lr}.
We incorporated the additional requirement for negation subject, and made corresponding adjustments w.r.t the Chinese expressions.  
In our guidelines, we require the semantic clarity and completeness of the labeled triplets, and the clear boundary of each element in a negation triplet. It should be noted that, as \cite{sfu_lr} we require that each element in a triplet is a continuous span in the sentence, and there should be no overlap between different elements. Such setting is more in line with the user reviews in Meituan's real-world scenario.

We recruited 8 workers and 2 experts to accomplish the annotation task, and adopted a annotation process with two stages as follows. In the first stage, the workers were provided with the sentences and the candidate cue list, and then annotated the negation triplets referring to the candidate cues. In the second stage, 
{for a batch of annotated data from the first stage}, the experts randomly selected $30\%$ of the annotated sentences, and validated them through correcting the incorrect annotations. Only batches with a validation accuracy more than 95\% were included into the final dataset. 
To measure the agreements between two annotators, we have calculated the Cohen's Kappa of NegComment, which achieves 89.72\%, indicating a high degree of consistency.

Finally, we obtained 7,049 validated sentences in NegComment, with a balanced distribution across the seven domains. 
For fine-tuning and evaluating the models, the ratio of training, validation and test set was set to 8:1:1.

\begin{table*}
  \centering
    \begin{tabular}{lrrrrrr}
    \toprule
    Model & \multicolumn{1}{l}{F1\%} & \multicolumn{1}{l}{impr.} & \multicolumn{1}{l}{P\%} & \multicolumn{1}{l}{impr.} & \multicolumn{1}{l}{R\%} & \multicolumn{1}{l}{impr.} \\
    \midrule
    NegBERT & 63.69  & 16.0\% & 62.81  & 17.2\% & 64.60  & 14.7\% \\
    BERT  & 63.91  & 15.7\% & 63.08  & 16.9\% & 64.74  & 14.5\% \\
    RoBERTa & 63.54  & 16.2\% &\underline{63.77}  & 16.0\% & 63.31  & 16.4\% \\
    XLNet & \underline{64.33}  & 15.1\% & 63.51  & 16.3\% & \underline{65.17}  & 13.9\% \\
    \midrule
    MultiHead & 63.90  & 15.7\% & 64.35  & 15.2\% & 63.44  & 16.2\% \\
    TPLinker & 64.92  & 14.3\% & 65.31  & 13.9\% & 64.54  & 14.7\% \\
    ETL-span & \underline{68.27}  & 9.9\% & \underline{68.88}  & 9.2\% & \underline{67.67}  & 10.6\% \\
    \midrule
    Span-ASTE & 67.92  & 10.4\% & 67.84  & 10.6\% & 68.01  & 10.1\% \\
    BART-ABSA & 59.63  & 21.3\% & 60.30  & 20.5\% & 58.97  & 22.1\% \\
    UIE   & \underline{72.21}  & 4.7\% & 71.15  & 6.2\% & \underline{73.30}  & 3.2\% \\
    Dual-MRC & 71.34  & 5.9\% & \underline{72.16}  & 4.9\% & 70.53  & 6.8\% \\
    \midrule
    ChatGPT (0 shot) & 24.82 & 67.2\% & 24.76 & 67.4\% & 24.88 & 67.1\% \\
    ChatGPT (5 shot) & 45.60  & 39.8\% & 45.52  & 40.0\% & 49.19  & 35.0\% \\
    ChatGPT (10 shot) & \underline{49.11}  & 35.2\% & \underline{49.03}  & 35.4\% & \underline{49.19}  & 35.0\% \\
    \midrule
    SSENE  & \textbf{75.78} &  & \textbf{75.88} &  & \textbf{75.69} & \\
    \bottomrule
    \end{tabular}
    \caption{All compared models' NTE performance on the NegComment dataset. 
    The best performance scores are bold and the best performance scores in each baseline group are underlined.
  For ChatGPT, we selected different numbers of samples for the setting of in-context learning. 
  }
  \label{tab:results}
\end{table*}

\subsection{Baselines}
As NTE is a new extraction task for which no previous models were designed, we selected models of relevant tasks as baselines, including CDSR, aspect-based sentiment analysis (ABSA), and entity and relation extraction (ERE). 
When using CDSR baselines, we additionally extract the negation subject based on the extracted cue and scope. 
For ERE models, we directly extract the negation subject and scope (regarded as the entity pair) on the premise that the cue is known, since these models cannot extract the cue span. 
For ABSA baselines, only generative models are considered given that they are more flexible to expand generating two spans to three span. 

In summary, we divide the baselines into three groups. 
(1) CDSR models. \textbf{NegBERT}~\cite{Khandelwal2020} proposes a 2-stage approach to extract cue and scope. \textbf{BERT/XLNet/RoBERTa}~\cite{khandelwal2020multitask} uses a multi-task learning framework, where BERT, XLNet \cite{yang2019xlnet} and RoBERTa \cite{liu2019roberta} are adopted as Encoder respectively. 
(2) ERE models. \textbf{MultiHead}~\cite{Bekoulis2018} identifies all candidate entities and performs relation extraction for each entity pair. \textbf{TPLinker}~\cite{Wang2021} employs a handshaking tagging strategy to perform the extraction task. \textbf{ETL-Span}~\cite{Yu2020} uses a decomposition strategy to split the joint extraction task and uses a sequence labeling model to extract the entities and relations. 
(3) Generative extraction models. \textbf{Dual-MRC}~\cite{mao2021joint} leverags two machine reading comprehension modules. \textbf{Span-ASTE}~\cite{Xu2021} uses a span-level approach. \textbf{BART-ASBA}~\cite{Yan2021} uses the BART model \cite{lewis2019bart} to perform extraction and classification. \textbf{UIE}~\cite{lu2022unified} proposes a structured extraction language to describe different information extraction tasks.

In addition, we also tested the performance of \textbf{ChatGPT}\footnote{\url{https://openai.com/blog/chatgpt}. The experiments were conducted using the version of ChatGPT prior to April 15, 2023.} on NTE with zero-shot, 5-shot, and 10-shot setting of in-context learning, since it has demonstrated its wonderful capabilities of natural language understanding. Our designed prompts for ChatGPT can be found in our source code.

\subsection{Implementation Setup}
We chose the Mengzi-T5\footnote{\url{https://github.com/Langboat/Mengzi}.} \cite{zhang2021mengzi} as the backbone model of our SSENE, {and used the package HanLP\footnote{\url{https://github.com/hankcs/HanLP}.} to get the dependency trees of sentences}. 
During the training of SSENE, we used the Adam optimizer with a learning rate of 2e-5 and a batch size of 12.
In addition, we conducted a grid search to optimize the hyperparameters of SSENE. 
{Based on our tuning studies, the loss controlling parameter $\alpha$ in Eq. \ref{eq:L} was set to 0.5, and the controlling parameters of transform function $f()$ in Eq. \ref{eq:M} were set to $\gamma_1=2, \gamma_2=0.5$.}

In all experiments, unless otherwise specified, all models were based on the T5-base architecture, which includes a total of 12 layers in both the encoder and decoder. Each transformer layer consists of 12 attention heads, with a hidden layer dimension of 768.

\subsection{Overall Performance Comparison}
In following experiments, we adopted precision (P), recall (R), and F1 scores as evaluation metrics. A triplet \emph{<subject, cue, scope>} is regarded as correct only if it matches the ground truth completely.

To answer RQ1, we display the overall results in Table ~\ref{tab:results}. 
Based on the results, we have the following analysis. 
(1) SSENE consistently outperforms all baselines, justifying the effectiveness of incorporating the sentence's syntactic information and the auxiliary task of semantic consistency on generating more accurate negation triplets. 
(2) For CDSR baselines in the first group, although subjects were additionally extracted based on the extracted cue, they are still not accurate enough since these baselines' F1 scores are both less than 65\%. 
(3) For ERE baselines, they still perform poorly though we only consider subject and scope for metric calculation, possibly due to that they tend to extract a term rather than a long span as the scope. 
(4) The generative baselines in the third group were designed specifically for different tasks, so the the performance discrepancy between them is more obvious than the other two groups. Among them, UIE has the best performance. 
(5) The performance of ChatGPT on NTE is poor compared with other models no matter what setting was adopted. We will investigate the reason in the subsequent case studies. 

\subsection{Ablation Study} \label{sec:ablation}
To answer RQ2, we further compared SSENE with its ablated variants, including SSENE-SD, SSENE-SC and SSENE-SD\&SC. In SSENE-SD, the sentence's syntactic dependency (SD) information is removed from encoder, i.e., the syntactic association matrix $\mathbf{M}$ is not used in Eq. \ref{eq:A}. In SSENE-SC, the auxiliary task learning of ensuring the semantic consistency (SC) between the negation triplet and sentence is removed. In SSENE-SD\&SC, both SD and SC are removed, simplifying it to be a fine-tuned T5-based seq2seq model. 

\begin{table}[t]
  \centering
  \begin{adjustbox}{max width=.5\textwidth}
    \begin{tabular}{lcccccc}
    \toprule
    Model & F1\% & {impr.} & {P\%} &{impr.} &{R\%} & {impr.} \\
    \midrule
    SSENE-SD & 73.14  & 3.6\% &  73.23  & 3.5\% &  73.05  & 3.5\% \\
    SSENE-SC &   \underline{74.31}  & 2.0\% &  \underline{74.40}  & 2.0\% &  \underline{74.22}  & 1.9\% \\
    SSENE-SD\&SC &          71.48  & 5.7\% &          71.57  & 5.7\% &    71.39  & 5.7\% \\
    \midrule
    SSENE  & \textbf{75.78} &  & \textbf{75.88} &  & \textbf{75.69} & \\
    \bottomrule
    \end{tabular}  
    \end{adjustbox}
    \caption{Ablation study results of SSENE on NegComment.}\label{tab:ab1}
\end{table}

Table \ref{tab:ab1} shows the NTE performance of SSENE and the three ablated variants, based on which we also have the following observations. 
At first, removing either SD or SC decreases the NTE performance of SSENE, indicating that both the syntactic information and the auxiliary task of semantic consistency are helpful to obtain more accurate NTE. 
In addition, compared with SC, SD is more helpful to improve NTE performance since SSENE's relative performance improvement over SSENE-SD is more than that over SSENE-SC. Hence, we further investigate the incorporation of syntactic information in the next experiments.

\begin{table}[t]
  \centering
  \begin{adjustbox}{max width=.5\textwidth}
    \begin{tabular}{lrrrrrr}
    \toprule
    Model & F1\% & {impr.} & {P\%} &{impr.} &{R\%} & {impr.} \\
    \midrule
    Abs. SPE & 68.95 & 7.8\% & 69.23  & 7.5\% & 68.66  & 8.1\% \\
    Rel. SPE & 70.29 & 5.7\% & 70.39 & 5.6\% & 70.19 & 5.7\% \\
    GATE  & \underline{72.59} & 2.4\% & \underline{72.68} & 2.4\% & \underline{72.49} & 2.4\% \\
    Syntax-BERT & 70.56 & 5.3\% & 70.46 & 5.6\% & 70.65 & 5.1\% \\
    SSENE-SC & \textbf{74.31} &  & \textbf{74.40} &  & \textbf{74.22} \\
    \bottomrule
    \end{tabular}
  \end{adjustbox}
  \caption{Comparisons of different syntactic information encoding methods.}
  \label{tab:encode}
\end{table}

\begin{table}[t]
  \centering
   \begin{adjustbox}{max width=.5\textwidth}
    \begin{tabular}{lrrrrrr}
    \toprule
    Model & F1\% & {impr.} & {P\%} &{impr.} &{R\%} & {impr.} \\
    \midrule
    Random & 65.35 & 13.7\% & 65.47 & 13.6\% & 65.23 & 13.8\% \\
    Noise (s=0.01) & \underline{72.59} & 4.0\% & \underline{72.68} & 4.1\% & \underline{72.49} & 4.0\% \\
    Noise (s=0.1) & 67.50 & 10.0\% & 67.60  & 10.1\%& 67.41  & 10.1\%\\
    SSENE-SC & \textbf{74.31} & & \textbf{74.40} & & \textbf{74.22} \\
    \bottomrule
    \end{tabular}
  \end{adjustbox}
  \caption{Comparisons of using different syntactic matrices in the sentence's encoder.}
  \label{tab:tree}
\end{table}

\subsection{Justifying the Way of Incorporating Syntactic Information}
To answer RQ3, we compared the effectiveness of the way encoding the sentence's syntactic dependency tree in SSENE, with the ways that were adopted by the previous works. 

Some existing works have also leveraged the syntactic dependency tree in Transformer. 
\cite{wang2019self} proposed a structure positional encoding (SPE) method based on dependency tree, including absolute SPE and relative SPE. 
\cite{ahmad2021gate} proposed a simple distance calculation method, namely GATE. Different to our method, GATE directly encodes the distance matrix as a 0-1 mask matrix for the self-attention's computation. \cite{bai2021improving} also converted the syntactic dependency tree into mask matrices, but manually designed the parent, child and sibling mask. 
Table~\ref{tab:encode} lists the results of these methods. For fair comparisons, we only compared them with SSENE-SC. It shows that the relative SPE is better than absolute SPE, but both are inferior to the rest three methods. Obviously, our SSENE has the best performance, justifying the superiority of our method of incorporating syntactic information.

Furthermore, we also investigated the significance of syntactic association matrix $\mathbf{M}$ through comparing the variants of disturbing $\mathbf{M}$'s value in $\mathbf{E}_{dep}$. In Table~\ref{tab:tree}, 
Random denotes the variant of replacing $\mathbf{M}$ with a randomly generated matrix. Noise (s=0.01/0.1) denotes the variants of introducing Gaussian random noise (with 0.01 and 0.1 variance) into the computation of $\mathbf{M}$. 
The results show that the $\mathbf{M}$ of incorporating the correct syntactic information indeed refines the self-attention's computation, resulting in more accurate NTE.

\begin{table*}[t]
  \centering
\scriptsize{
    \begin{tabular}{p{0.08\linewidth}p{0.88\linewidth}}
    \toprule
    Sentence & 很糟心的游玩体验。游乐场的游乐设施很多都不开放，有的还不能用套票，真的是不划算。A very bad experience. Many rides at the amusement park are not open, and some of them are not available in package, which is really not good value. \\
    \midrule
    NegBERT  & <游乐设施rides, 不not, 开放open>, <游乐设施rides, 不能not, \underline{套票package}>, <\underline{游乐设施rides}, 不not, 划算good value> \\
    \midrule
    ETL-span  & <游乐设施rides, -, 开放open>, <游乐设施rides, -, \underline{套票package}>, <游乐场amusement park, -, 划算good value> \\
    \midrule
    UIE & <游乐设施rides, 不not, 开放open>, <\underline{有的some}, 不能not, 用套票available in package>, <游乐场amusement park, 不not, 划算good value> \\
    \midrule
    ChatGPT (10 shot)  & <\underline{游玩体验experience, 糟bad, 心bad}>, <游乐设施rides, 不not, 开放open>, <\underline{有的some}, 不能not, 用套票available in package>, <游乐场amusement park, 不not, 划算good value> \\
    \midrule
    SSENE & <游乐设施rides, 不not, 开放open>, <游乐设施rides, 不能not, 用套票available in package>, <游乐场amusement park, 不not, 划算good value> \\
    \midrule
    Ground Truth  & <游乐设施rides, 不not, 开放open>, <游乐设施rides, 不能not, 用套票available in package>, <游乐场amusement park, 不not, 划算good value> \\
    \bottomrule
    \end{tabular}
  \caption{A case's results. The Underlines indicate the incorrect extracted elements. }\label{tab:case}  
  }
\end{table*}

\begin{table*}[t]
\centering
\scriptsize{
    \begin{tabular}{p{0.08\linewidth}p{0.88\linewidth}}
    \toprule
    Sentence & 酒店的前台态度不好，服务也不周到，效率低下，感觉很不专业，不符合星级酒店的标准。 The attitude of the hotel's receptionist is not good, the service is also not attentive, the efficiency is low, feeling very unprofessional, does not meet a star hotel's standard.  \\
    \midrule
   SSENE-SD & <态度attitude, 不not, 好good>, <服务service, 不not, 周到attentive>, <\underline{服务service}, 不un, 专业professional>, <\underline{星级酒店star hotel, 不not, 符合meet}> \\
    \midrule
   SSENE-SC & <态度attitude, 不not, 好good>, <服务service, 不not, 周到attentive>, <前台receptionist, 不un, 专业professional>, <\underline{酒店hotel}, 不not, 符合星级酒店的标准meet a star hotel's standards> \\
    \midrule
    SSENE-SD\&SC & <态度attitude, 不not, 好good>, <服务service, 不not, 周到attentive>, <\underline{感觉feeling}, 不un, 专业professional>, <\underline{星级酒店star hotel, 不not, 符合meet}> \\
    \midrule
    SSENE  & <态度attitude, 不not, 好good>, <服务service, 不not, 周到attentive>, <前台receptionist, 不un, 专业professional>, <前台receptionist, 不not, 符合星级酒店的标准meet a star hotel's standard> \\
    \midrule
    Ground Truth  & <态度attitude, 不not, 好good>, <服务service, 不not, 周到attentive>, <前台receptionist, 不un, 专业professional, <前台receptionist, 不not, 符合星级酒店的标准meet a star hotel's standards> \\
    \bottomrule
    \end{tabular}
     \caption{A case's extraction results of SSENE and its ablated variants.}\label{tb:case}
  \label{tab:ab2}
  }
\end{table*}

\subsection{Justifying the Impact of Syntactic Dependency Attention}
\begin{figure*}[t]
    \centering
    \includegraphics[width=\textwidth]{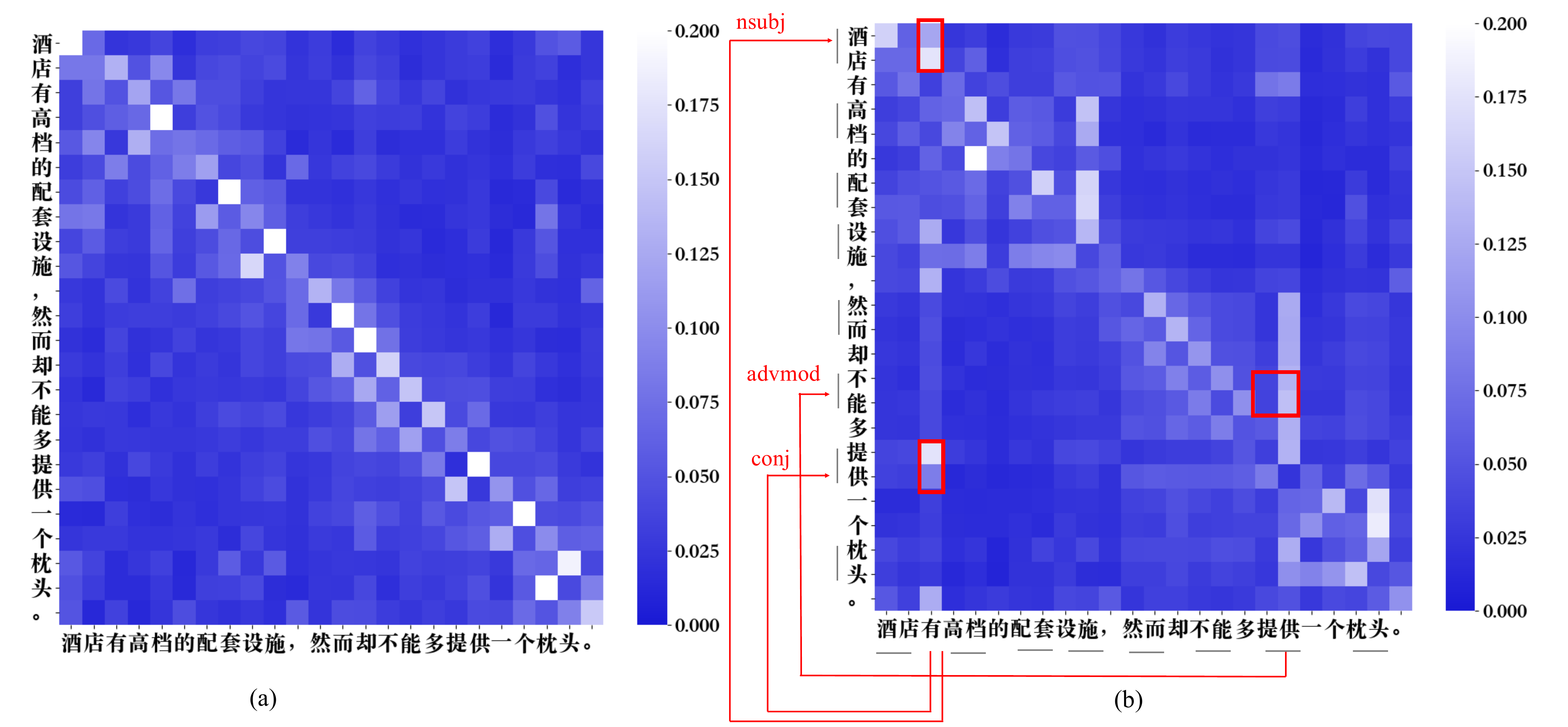}
    \caption{For the sentence ``酒店有高档的配套设施，然而却不能多提供一个枕头。|The hotel has luxurious facilities, yet cannot offer one more pillow.'', Subfigure (a) and (b) display the average attention distribution of the first layer in SSENE's Encoder before and after incorporating the syntactic association matrix $\mathbf{M}$, respectively. 
    }
    \label{fig:att}
\end{figure*}

To investigate the impact of introducing the syntactic association matrix $\mathbf{M}$ on the self-attention's computation in SSENE's syntax-aware encoder, we visualize the attention distributions in $\mathbf{E}_{dep}$ before and after injecting $\mathbf{M}$, as shown in Fig. \ref{fig:att}(a) and (b). In the figure, the grey lines indicate the tokenization results, and the red arrows indicate some important long-distance dependency relationships between the terms that come from the sentence's syntactic dependency tree. And the grids of high attention values are marked with light color. 

It shows that the incorporation of syntactic dependency information increases the influence of long-distance attentions. For example, for extracting the correct negation triplet <酒店hotel，不能cannot, 多提供一个枕头offer one more pillow>, the distant subject is correlated to the cue and scope through the syntactic dependency conjunctions of ``有has'' and ``提供provide'', which are highlighted by the attentions refined by $\mathbf{M}$, as shown Fig. \ref{fig:att} (b).

\subsection{Case Study}
We further investigated our SSENE's advantage through case studies. Table \ref{tab:case} displays the extraction results of SSENE with representative baselines, where three negation triplets can be extracted as shown by Ground Truth. 
It shows that NegBERT incorrectly identifies ``游乐设施rides'' as the negation subject of cue ``不not'' and scope ``划算good value'', indicating weak capabilities in recognizing subjects for CDSR models.
ETL-span mistakenly extracts ``套票package'' as negation scope, because this baseline was proposed specifically for entity-pair extraction, thus tending  to extract the entity's name instead of the phrase ``用套票available in package''. 
In addition, UIE mistakenly extracted ``有的some'' as the subject of ``不能not'', implying that it tends to regard the term close to the cue as the subject instead of the distant term. 
Comparatively, our SSENE can extract the correct subject ``游乐设施rides'' with the aid of syntactic dependency although it is distant. 
In the extraction results of ChatGPT, <游玩体验experience, 糟bad, 心bad> is not a correct negation triplet but an expression of negative sentiment. It implies that ChatGPT misunderstands the negation as the expression of negative sentiment rather than from the perspective of syntax. 

We also display another instance's extraction results generated by SSENE and its ablated variants in Table \ref{tab:ab2}. The results show that the triplet <星级酒店star hotel, 不not, 符合meet> of incorrect syntax is generated if the syntactic dependency (SD) information is not incorporated into SSENE. Furthermore, the correct subject ``前台receptionist'' can not generated as the cue ``不un'' and scope ``专业professional'', while the wrong subject ``服务service'' and ``感觉feeling'' are generated. In addition, without considering the semantic consistency (SC), the incorrect triplet  <酒店hotel, 不not, 符合星级酒店的标准meet a star hotel's standards> is generated which is not semantically consistent with the sentence in fact.

\subsection{Validation on Negation CDSR}
Although our SSENE is devised to achieve NTE, the semantic consistency between the sentence and the extraction result, as well as the aim of the syntax-aware encoder are also effective on other sentence-level extraction tasks. 
To verify this (RQ4), we further tested the performance of SSENE on Negation CDSR, upon the English benchmark SFU Review dataset ~\cite{sfu_lr}. Similar to our setup, this dataset imposes constraints where the negation cue is not included by the scope, and both of them are not separate spans. 

To achieve this task, for the ERE models, we fixed the relation types and used them to extract cue-scope pairs. For the generative models, we adapted their data format to meet CDSR. 
The results in Table \ref{tab:cdsr} show that SSENE still obtains the best performance on this CDSR task.

\begin{table}[t]
  \centering
  \begin{adjustbox}{max width=.5\textwidth}
    \begin{tabular}{lrrrrrr}
    \toprule
    Model & F1\% & {impr.} & {P\%} &{impr.} &{R\%} & {impr.} \\
    \midrule
    NegBERT & 72.27 & 3.5\% & 72.15 & 4.0\% & 72.38 & 3.0\% \\
    BERT  & 71.24 & 5.0\% & 71.01 & 5.6\% & 71.48 & 4.3\% \\
    RoBERTa & 73.27 & 2.1\% & 73.50 & 2.0\% & 73.04 & 2.1\% \\
    XLNet & \underline{73.52} & 1.7\% & \underline{73.75} & 1.7\% & \underline{73.29} & 1.7\% \\
    \midrule
    MultiHead  & 64.78 & 15.4\% & 65.00 & 15.4\% & 64.57 & 15.5\% \\
    TPLinker & 66.98 & 11.6\% & 67.19 & 11.6\% & 66.77 & 11.7\% \\
    ETL-Span & \underline{68.91} & 8.5\% & \underline{69.12} & 8.5\% & \underline{68.71} & 8.5\% \\
    \midrule
    Span-ASTE & 67.01 & 11.6\% & 67.24 & 11.5\% & 66.78 & 11.7\% \\
    BART-ABSA & 60.09 & 24.4\% & 60.00 & 25.0\% & 60.19 & 23.9\% \\
    UIE   & \underline{74.02} & 1.0\% & \underline{74.24} & 1.0\% & \underline{73.80} & 1.0\% \\
    Dual-MRC & 70.28 & 6.4\% & 70.50 & 6.4\% & 70.06 & 6.4\% \\
    \midrule
    ChatGPT(0) & 63.58 & 17.6\% & 61.62 & 21.7\% & 65.67 & 13.5\% \\
    ChatGPT(5) & 67.87 & 10.2\% & 66.18 & 13.3\% & 69.66 & 7.0\% \\
    ChatGPT(10) & \underline{68.45} & 9.2\% & \underline{66.28} & 13.2\% & \underline{70.77} & 5.4\% \\
    \midrule
    SSENE & \textbf{74.78} &       & \textbf{75.00} &       & \textbf{74.56} &  \\
    \bottomrule
    \end{tabular}
  \end{adjustbox}
  \caption{All compared models’ CDSR performance.}
  \label{tab:cdsr}
\end{table}

\section{Related Work}
\paragraph{Joint Entity and Relation Extraction}
Many works have focused on joint entity and relation extraction (ERE). \cite{Zheng2017} proposes a tagging scheme that combines the entity role and relation, transforming ERE task into sequence labeling. To address overlaps, \cite{Bekoulis2018} extracts candidate entities and predicts relations for each entity pair. \cite{Nayak2020,ren2022simple} uses an encoder-decoder architecture with a pointer network for triplet generation. \cite{Wei2020} introduces the \emph{CasRel} framework to handle overlapping triplets. It identifies subjects first and then uses span-based tagging to identify corresponding objects based on easy relations. Wang et al. \cite{Wang2021} proposed \emph{TPLinker}, which is a one-stage method converting the task into token pair linking. Their handshaking tagging scheme aligns the boundary tokens of entity pairs under each relation type.

\paragraph{Negation Cue Detection and Scope Resolution}
The study of negation understanding initially emerged from the analysis of biomedical texts, where discerning the presence or absence of specific characteristics is crucial for patient care. Researchers in the pertinent fields have structured negation understanding into two distinct sub-tasks: cue detection, which involves identifying triggers of negation, and scope resolution, which entails determining the spans of text affected by the negation. 
Traditionally, some rule-based systems like NegEx~\cite{chapman2001simple} were popular, which used regular expressions. However, they struggled to be generalized to other domains. Afterwards, the approaches of supervised learning models were proposed, wherein the common solutions include feature extraction and token-level classification. For example, \cite{lazib2016negation} employs an RNN-based architecture, while \cite{chen2019attention} combines bidirectional LSTM with attention mechanism.
Recently, NegBERT~\cite{Khandelwal2020} utilizes BERT for negation detection, and \cite{truong2022improving} introduced an augmenting approach that emphasizes negation and extends the transfer learning capability of NegBERT through additional pretraining tasks.

\paragraph{Negation in Language Models}
PLMs have demonstrated excellent performance in various natural language understanding and generation tasks. However, they still struggle with capturing negation semantics. \cite{Kassner2020} finds that PLMs would generate both factual statements (e.g., ``Birds can fly'') and their incorrect negations, indicating the lack of understanding negation semantics. \cite{hossain2020analysis} have investigated the role of negation in natural language inference and argued that existing benchmarks fail to express its semantic role accurately. 
To enhance syntactic abilities, \cite{Noji2020} and \cite{Min2020} leverage negative examples and employ syntactic data augmentation techniques, respectively. 
\cite{Hosseini2021} proposed the unlikelihood objective to fine-tune LMs and used sentence dependencies, part-of-speech tags, and morphological features to deterministically generate negated versions of original sentences.

\section{Conclusion}
To address the problem of missing the negation subject in traditional negation understanding, we propose a new task of negation triplet extraction in this paper and propose a novel model SSENE to achieve this new task. Our SSENE is built based on a generative PLM with a multi-task learning framework. We specially devise a syntax-aware encoder in SSENE to incorporate the input sentence's syntactic dependency information, and ensure the semantic consistency between the extracted negation triplets and the sentence through the auxiliary task learning. 
Our extensive experiments upon the dataset constructed from Meituan's real-world scenarios demonstrate our SSENE's superiority over the baselines. Furthermore, the impacts of incorporating the syntactic dependency information and semantic consistency are also justified.

\end{CJK}

\section*{Limitations}
In this paper, due to time constraints, the effectiveness of the proposed SEENE model was only validated on the Negation CDSR task. Future research could explore the efficacy of the syntax-aware encoder and semantic consistency in other relation extraction tasks.

\section*{Acknowledgements}
This work was supported by the Chinese NSF Major Research Plan (No. 92270121), National Natural Science Foundation of China (No. 62306112), Shanghai Science and Technology Innovation Action Plan (No.21511100401), Sailing Program (No. 23YF1409400). The computations in this research were performed using the CFFF platform of Fudan University.

\section*{Bibliographical References} \label{sec:reference}
\bibliographystyle{lrec-coling2024-natbib}
\bibliography{main}



\end{document}